\documentclass[letterpaper]{article} 
\usepackage{aaai24}  
\usepackage{times}  
\usepackage{helvet}  
\usepackage{courier}  
\usepackage[hyphens]{url}  
\usepackage{graphicx} 
\usepackage{natbib}  
\usepackage{caption} 
\usepackage[linesnumbered, ruled, vlined]{algorithm2e}
\SetArgSty{textrm}
\SetInd{0.1em}{0.5em}
\SetCommentSty{emph}
\SetAlgoSkip{}
\usepackage{etoolbox}
\makeatletter
\patchcmd{\@algocf@start}
  {-1.5em}
  {0pt}
  {}{}
\makeatother

\usepackage{amssymb}
\usepackage{amsmath}
\usepackage{bm}
\usepackage{multirow}

\DeclareMathOperator*{\argsort}{arg\,sort}
\DeclareMathOperator*{\argmin}{arg\,min}

\usepackage{amsthm}
\makeatletter
\def\thm@space@setup{%
 \thm@preskip=\parskip \thm@postskip=0pt
}
\makeatother
\usepackage{thmtools}
\declaretheorem{theorem}
\declaretheoremstyle[%
  spaceabove=0pt,
  spacebelow=0pt,
  headfont=\normalfont\itshape,%
  postheadspace=1em,%
  qed=\qedsymbol%
]{mystyle}
\declaretheorem[name={Proof},style=mystyle,unnumbered,
]{myproof}

\title{Large-Scale Multi-Robot Coverage Path Planning via Local Search\thanks{Code: \protect\url{https://github.com/reso1/LS-MCPP}}}
\author{
    Jingtao Tang, Hang Ma
}
\affiliations{
    Simon Fraser University\\

    \{jingtao\_tang, hangma\}@sfu.ca
}

\begin{document}

\maketitle

\begin{abstract}
We study graph-based Multi-Robot Coverage Path Planning (MCPP) that aims to compute coverage paths for multiple robots to cover all vertices of a given 2D grid terrain graph $G$. Existing graph-based MCPP algorithms first compute a tree cover on $G$---a forest of multiple trees that cover all vertices---and then employ the Spanning Tree Coverage (STC) paradigm to generate coverage paths on the decomposed graph $D$ of the terrain graph $G$ by circumnavigating the edges of the computed trees, aiming to optimize the makespan (i.e., the maximum coverage path cost among all robots).
In this paper, we take a different approach by exploring how to systematically search for good coverage paths directly on $D$. We introduce a new algorithmic framework, called LS-MCPP, which leverages a local search to operate directly on $D$. We propose a novel standalone paradigm, Extended-STC (ESTC), that extends STC to achieve complete coverage for MCPP on any decomposed graphs, even those resulting from incomplete terrain graphs. Furthermore, we demonstrate how to integrate ESTC with three novel types of neighborhood operators into our framework to effectively guide its search process. Our extensive experiments demonstrate the effectiveness of LS-MCPP, consistently improving the initial solution returned by two state-of-the-art baseline algorithms that compute suboptimal tree covers on $G$, with a notable reduction in makespan by up to 35.7\% and 30.3\%, respectively. Moreover, LS-MCPP consistently matches or surpasses the results of optimal tree cover computation, achieving these outcomes with orders of magnitude faster runtime, thereby showcasing its significant benefits for large-scale real-world coverage tasks.
\end{abstract}

\section{Introduction}
Coverage path planning (CPP) is a fundamental problem~\cite{galceran2013survey} in robotics, which aims to find a path for a robot to completely cover a terrain of interest, such as indoor floors~\cite{bormann2018indoor} and outdoor fields~\cite{torres2016coverage}.
Multi-Robot Coverage Path Planning (MCPP) is an extension of CPP tailored for multi-robot systems, aiming to coordinate the paths of multiple robots to completely cover the given terrain.
With improved task efficiency and system robustness, MCPP has facilitated diverse real-world applications, including environmental monitoring~\cite{collins2021scalable} and search and rescue~\cite{song2022multi}.
A fundamental challenge of MCPP lies in generating cost-balancing coverage paths to optimize task efficiency, commonly quantified by the \textit{makespan}, which is the maximum path cost of all robots.
This challenge is further compounded when dealing with large-scale applications where the number of robots and the size of the terrain increase.
\begin{figure}[t]
\centering
\includegraphics[width=1\columnwidth]{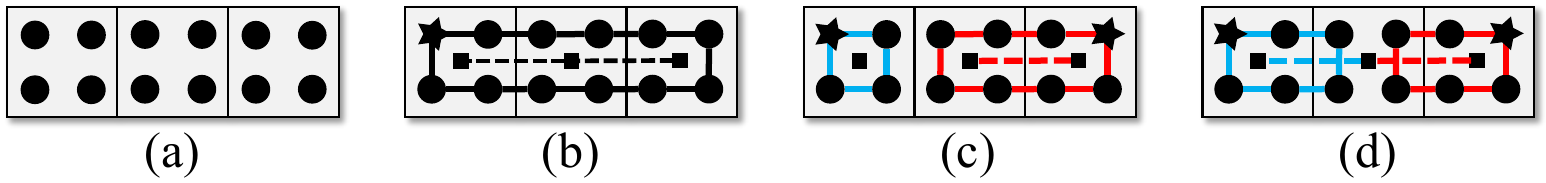} 
\caption{Graph-based CPP and MCPP: Gray squares, black circles, and black stars represent terrain graph vertices, decomposed graph vertices, and initial vertices of robots, respectively; Solid lines and dashed lines represent coverage paths and spanning edges, respectively. (a) Terrain graph with uniform edge weights. (b) The single-robot coverage path generated by STC. (c)(d) Suboptimal and optimal $2$-robot coverage paths with makespans $2$ and $1.5$, respectively.}
\label{fig:STC}
\end{figure}

In this paper, we follow existing graph-based MCPP algorithms~\cite{zheng2010multirobot,li2023sp2e} that represent the terrain to be covered as a 4-connected 2D grid graph $G$, where each edge connects horizontally or vertically adjacent vertices. The robots are required to start at and return to their respective initial vertices, as in the \textit{cover and return} setting~\cite{zheng2007robot}.
The foundation of these graph-based MCPP algorithms lies in the Spanning Tree Coverage (STC) paradigm~\cite{gabriely2001spanning, gabriely2002spiral}, initially developed for (single-robot) CPP. STC operates on the terrain graph $G$ but finds a coverage path with minimal makespan on the decomposed graph $D$ derived from $G$. The decomposed graph $D$ is also a 4-connected 2D grid graph, resulting from decomposing each vertex of $G$ into four decomposed vertices. Fig.~\ref{fig:STC} shows the terrain graph $G$ and its corresponding decomposed graph $D$ of an example terrain to be covered, where STC generates a single-robot coverage path on $D$ by circumnavigating (i.e., always moving along the right side of the spanning edges) the minimum spanning tree of $G$.

Like STC, existing graph-based MCPP algorithms operate on the given terrain graph $G$ exclusively to build a tree cover---a forest of multiple trees, each rooted at the initial vertex of a robot, that jointly cover all vertices of $G$. The coverage path for each robot is then obtained by circumnavigating its corresponding tree. In essence, these algorithms reduce MCPP to the NP-hard Min-Max Rooted Tree Cover problem \cite{even2004min, nagamochi2007approximating} on $G$ that aims to optimize the weight of the largest-weighted tree in the tree cover since it determines the makespan of the resulting coverage paths on $D$. However, operating on the terrain graph $G$ exclusively has two disadvantages. Firstly, it does not work for an incomplete terrain graph $G$ where some of the four decomposed vertices of a vertex might be blocked and thus absent in the decomposed graph $D$. Secondly, an optimal tree cover on $G$ does not necessarily result in an optimal MCPP solution (as illustrated in Fig.~\ref{fig:STC}-(c) and (d)), which yields an asymptotic suboptimality ratio of four for makespan in the worst case~\cite{zheng2010multirobot}, since circumnavigating the trees in a tree cover explores only a portion of the solution space that encompasses all possible sets of coverage paths on the decomposed graph $D$.

Therefore, we take a different route to explore how to systematically search for good coverage paths directly on the decomposed graph. Our algorithmic contribution is through the following key aspects:
(1) We propose a novel standalone algorithmic paradigm called Extended-STC (ESTC), an extension of STC, to address coverage planning problems on both complete and incomplete terrain graphs by directly operating on decomposed graphs. Importantly, we demonstrate that ESTC guarantees complete coverage for both single- and multi-robot settings, rendering it an efficient and versatile solution to coverage path planning.
(2) We propose three types of specialized neighborhood operators to facilitate an effective local search process by identifying cost-efficient subgraphs of the decomposed graph that are used to generate coverage paths for the robots. The strategic integration of these operators enhances the efficiency of the local search in exploring the solution space.
(3) We demonstrate how to combine these neighborhood operators with iterative calls to the ESTC paradigm to establish our proposed LS-MCPP framework for solving MCPP.
To validate the effectiveness of LS-MCPP, we conduct extensive experiments, comparing it with three state-of-the-art baseline graph-based MCPP algorithms that operate on complete terrain graphs only. The results show that LS-MCPP achieves makespans that are smaller by up to 35.7\% and 30.3\% than two of the baseline algorithms, respectively, which compute suboptimal tree covers on the terrain graph. Additionally, LS-MCPP consistently achieves makespans comparable to or better than those achieved by the remaining baseline algorithm, which employs mixed integer programming (MIP) to compute optimal tree covers on the terrain graph. While the baseline algorithm takes more than tens of hours to complete, LS-MCPP accomplishes the same task in just a matter of minutes, showcasing its efficiency and practicality for large-scale real-world coverage problems.

\section{Related Work on MCPP}
Existing MCPP algorithms first partition the given terrain into multiple regions and then use a single-robot CPP algorithm to generate a coverage path on each region for a robot.
Without loss of generality, we categorize existing MCPP algorithms into decomposition-based and graph-based.
Decomposition-based algorithms first partition the terrain using geometric critical points found on the polylines of the terrain to be covered~\cite{karapetyan2017efficient,vandermeulen2019turn} and then generate zigzag paths covering each partitioned region.
Despite their simplicity in relatively open environments, they are unsuitable for obstacle-rich terrains (e.g., mazes) due to their reliance on geometric partitioning and cannot consider non-uniform traversal costs for weighted terrains.

Our paper focuses on graph-based MCPP algorithms~\cite{zheng2010multirobot, li2023sp2e} that are more versatile as they model the terrain to be covered as a graph and account for non-uniform traversal costs by assigning weights to the edges or vertices.
Solving MCPP on a graph for makespan minimization has been proved to be NP-hard~\cite{zheng2010multirobot}. Existing approaches include polynomial-time approximation algorithms~\cite{hazon2005redundancy, tang2021mstc}, numerical optimization~\cite{kapoutsis2017darp}, and an MIP formulation~\cite{tang2023mixed}.
A closely related problem is the multi-traveling salesman problem (mTSP) with a minmax objective~\cite{francca1995m}, which aims to find optimal routes for multiple salesmen to visit all given cities. However, existing scalable mTSP algorithms are designed for Euclidean spaces~\cite{he2023memetic}. We are unaware of any mTSP algorithms directly applicable to the graph-based MCPP problem.

\section{Problem Formulation}
In an MCPP instance, we are given a connected 2D grid terrain graph $G=(V_g, E_g)$ and its corresponding decomposed graph $D=(V_d, E_d)$, where each terrain vertex in $V_g$ is decomposed into four small adjacent vertices in $V_d$ as shown in Fig.~\ref{fig:STC}-(a). For clarity, we let $\delta_v\in V_g$ denote the corresponding terrain vertex of a decomposed $v\in V_d$ and $\varepsilon=(\delta_u, \delta_v)\in E_g$ denote the edge connecting $\delta_u$ and $\delta_v$. 
Each $\varepsilon\in E_g$ has a weight of $w_\varepsilon$.
Note that the definition of weighted edges is different from the literature that considers weighted vertices~\cite{zheng2010multirobot}, yet it is rather convenient to convert weighted edges to weighted vertices by evenly distributing each edge weight to its two endpoint vertices.
In that case, we have $w_{\delta_v}=\sum_{\varepsilon\sim {\delta_v}}w_\varepsilon$ for each $\delta_v\in V_g$, where $\varepsilon\sim \delta_v$ represents that $\delta_v$ is an endpoint of $\varepsilon$.
Each edge $e=(u,v)\in E_d$ of $D$ is thereby defined to have a weight $w_e=\frac{1}{4}\cdot\frac{1}{2}\cdot\left[\sum_{\varepsilon\sim\delta_u}w_{\varepsilon}+\sum_{\varepsilon\sim\delta_v}w_{\varepsilon}\right]$ by evenly splitting the weight $w_\delta$ of each $\delta\in V_g$ into $w_\delta/4$ for each of its four decomposed vertices~\cite{tang2023mixed}.

For any path $\pi=(e_1,e_2,...,e_{|\pi|})$ on $D$ where each $e_i\in E_{d}$, we define its cost as $c(\pi)=\sum_{i=1}^{|\pi|}w_{e_i}$.
Given a set $I=\{1,2,...,k\}$ robots with a set $R=\{r_i\}_{i\in I}\subseteq V_d$ of initial root vertices, the graph-based MCPP problem is to find a set $\Pi=\{\pi_i\}_{i\in I}$ of $k$ paths such that each $v\in V_d$ is visited by at least one coverage path for complete coverage and each $\pi_i$ starts and ends at $r_i$. The solution quality is measured by the makespan $\tau = \max\{c(\pi_1), c(\pi_2), ..., c(\pi_k)\}$.

\noindent\textbf{Notation:}
We use the tuple $(G, D, R)$ to denote an MCPP instance.
For each robot $i\in I$, we differentiate its corresponding subgraph $D_i=(V_{d,i}, E_{d,i})$ of $D$ 
and its corresponding subgraph $G_i=(V_{g,i}, E_{g,i})$ of $G$.

\section{Extended-STC}\label{sec:estc}
In this section, we present the new Extended-STC (ESTC) paradigm to address coverage problems on incomplete terrain graphs. We define that $\delta_v\in V_g$ is \textit{complete} if all its four decomposed vertices are present in $V_d$ and \textit{incomplete} otherwise. A terrain graph $G$ is \textit{complete} if all its vertices are complete and \textit{incomplete} otherwise. STC works only for complete terrain graphs due to its reliance on circumnavigating spanning edges of $G$. As in Fig.~\ref{fig:STC}-(c)(d), this restriction of STC also prevents STC-based MCPP algorithms~\cite{tang2023mixed} from finding good solutions even on complete terrain graphs. In this paper, we focus on covering a connected but possibly incomplete decomposed graph, as covering an unconnected decomposed graph only additionally requires a robot in each of its connected components.

\begin{figure}[t]
\centering
\includegraphics[width=0.85\columnwidth]{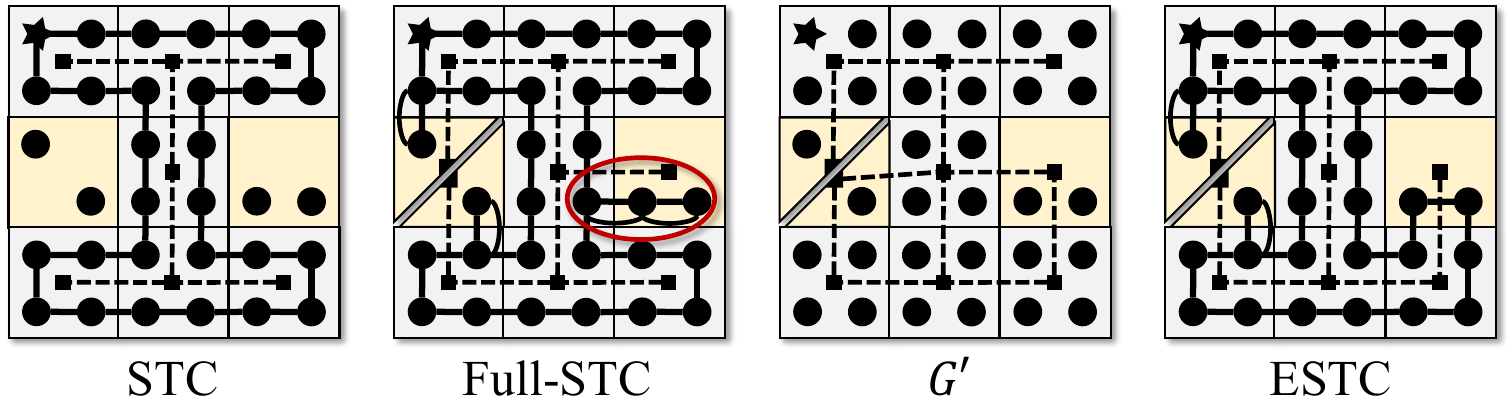} 
\caption{A CPP instance with a root vertex (black star) and incomplete terrain vertices (yellow cells), where STC has no complete coverage. Full-STC is sub-optimal with undesirable routes (red circled area), and ESTC is optimal based on the augmented terrain graph $G'$.}
\label{fig:ESTC}
\end{figure}

In order to generate coverage paths through incomplete terrain vertices, our proposed ESTC paradigm cleverly integrates a path-deformation procedure of Full-STC~\cite{gabriely2002spiral} into the offline computation of STC. Full-STC is specifically designed for online (single-robot) CPP on an unknown terrain graph. It can be applied to incomplete terrain graphs but is suboptimal for CPP instances on known terrain graphs since it does not consider different path costs resulting from different spanning trees.
Fig.~\ref{fig:ESTC} shows an example of the suboptimality of Full-STC.

Like Full-STC, ESTC operates on the augmented terrain graph $G'$ where some edges in $G$ are removed in $G'$ to reflect the connectivity between its terrain vertices. Specifically, edge $\varepsilon=(\delta_u, \delta_v)\in E_g$ is removed in $G'$ if each decomposed vertex $u$ of terrain vertex $\delta_u$ is nonadjacent to each decomposed vertex $v$ of terrain vertex $\delta_v$. A special case is that each incomplete terrain vertex $\delta\in V_g$ with only two diagonally opposite decomposed vertices present is replaced with two nonadjacent terrain vertices in $G'$ since its two decomposed vertices are nonadjacent.
Unlike Full-STC, ESTC considers non-uniform edge weights.
An edge $\varepsilon=(\delta_u, \delta_v)$ has the same weight as in $G$ if both $\delta_u$ and $\delta_v$ are complete. Otherwise, $\varepsilon$ has a manipulated edge weight of
\begin{align}\label{eqn:new_edge_weight}
w_{\max}\cdot\frac{1}{2}\cdot\left({\textstyle\sum_{\varepsilon\sim\delta_u}w_{\varepsilon}+\sum_{\varepsilon\sim\delta_v} w_{\varepsilon}}\right)
\end{align}
where $w_{\max}$ is the maximal edge weights of $E_g$ to prioritize using edges connecting complete terrain vertices.
Fig.~\ref{fig:ESTC} shows an example of the augmented terrain graph $G'$.

By construction, $G'$ is connected for a connected $D$. ESTC then builds a minimum spanning tree (MST) on $G'$ instead of $G$ and uses the same path-deformation rule defined as in Full-STC~\cite{gabriely2002spiral} to route through incomplete vertices. In essence, this rule reroutes the coverage path on the left side of a spanning edge whenever moving along the right side is impossible due to nonadjacency between decomposed vertices. 
The manipulated edge weight in Eqn.~(\ref{eqn:new_edge_weight}) prioritizes making each incomplete vertex a leaf of the MST, which prevents some decomposed vertices from repeatedly covering due to rerouting in path deformation and can thus improve the CPP solution quality (see Fig.~\ref{fig:ESTC}).
ESTC is guaranteed to generate a path that covers all vertices of a connected decomposed graph $D$ based on the argument used in the proof of Lemma 3.1 in~\cite{gabriely2002spiral} since the argument is not affected by the addition of edge weights and prioritization of edges.


\noindent\textbf{Replacing STC with ESTC for MCPP:}
Recall that for an MCPP instance $(G, D, R)$, STC-based MCPP algorithms first build $k$ trees of $G$ and then apply STC to generate the coverage paths. Therefore, they are restricted to generating each coverage path only on terrain graphs with complete vertices, which is suboptimal for most cases.
We propose to extend ESTC to MCPP as follows. First, we generate a set $\{D_i=(V_{d,i},E_{d,i})\}_{i\in I}$ of $k$ connected subgraphs of $D$. Each $D_i$ induces a subgraph $G_i$ of $G$ that is possibly incomplete. Then, we solve each sub-CPP instance ($G_i$, $D_i$, $r_i$) as described above by constructing an MST on the augmented $G_i'$ to generate the coverage path for each robot (see Fig.~\ref{fig:STC}-(d) for an example). The following theorem shows a sufficient condition for complete coverage.
\begin{theorem}
If $D_i$ is connected and $\bigcup_{i\in I} V_{d,i}=V_d$ for all $i$, ESTC is guaranteed to achieve complete coverage.
\end{theorem}
\begin{myproof}
Given that $D_i$ is connected, ESTC is guaranteed to generate a path that covers all its vertices, analogous to the argument used in the proof of Lemma 3.1 in~\cite{gabriely2002spiral}. Since $\bigcup_{i\in I} V_{d,i}=V_d$, the coverage paths generated for all $D_i$ jointly cover all vertices of $D$.
\end{myproof}

The design of our LS-MCPP framework is based on the principle expounded in the above theorem. It aims to search for a good set of subgraphs $D_i$ for the robots, each being connected, with the property that the union of all their vertices is the set of vertices $D$ to ensure complete coverage.


\section{Boundary Editing Operators}\label{sec:operators}
In this section, we introduce three types of boundary editing operators: grow operators, deduplicate operators, and exchange operators. These operators are designed to modify the boundaries of each subgraph $D_i$, involving the addition or removal of vertices.
We define the \textit{duplication} set $V^+=\{v\in V_d\,|\,n_v>1 \}$ as the set of decomposed vertices covered by more than one robot, where $n_v=\sum_{i\in I}\left|\{x\in V_{d,i}|x=v\}\right|$ counts the occurrences of vertex $v\in V_d$ across all subgraphs.
A boundary editing operator alters the set $\{D_i\}_{i\in I}$ of subgraphs but ensures that its property of complete coverage (i.e., $\bigcup_{i\in I}V_{d,i}=V_d$) and subgraph connectivity remains invariant. The operator aims to introduce only necessary duplication, which holds the potential to improve the resulting MCPP solution on the updated set of subgraphs with ESTC.

To allow for an efficient transition between different sets of subgraphs, we restrict our operator design to be edge-wise on any subgraph; that is, each operator involves the addition or removal of only one edge for each subgraph. This design is based on the observation that manipulating individual vertices can sometimes introduce unnecessary duplication to the coverage paths when ESTC is applied to the resulting subgraphs, as is exemplified by undesirable routes from a vertex $u\in V_{d,i}$ to its neighbor $v\in V_{d,i}$ and then back to $u$. In contrast, edge-wise operators modify two vertices simultaneously during each operation, eliminating such concerns and ensuring a more effective adjustment. Remarkably, edge-wise operators can achieve the same outcome as operators involving multiple edges. 
In addition, our operators add or remove only edges whose two endpoints are decomposed from the same terrain vertex since manipulating other edges would introduce unnecessary duplication.
Formally, we denote the set of such edges of any $D_i$ as
\begin{align}
F_{i}=\{(u,v)\in E_{d,i}\,|\,\delta_u=\delta_v\}.
\end{align}

\begin{figure}[t]
\centering
\includegraphics[width=0.85\columnwidth]{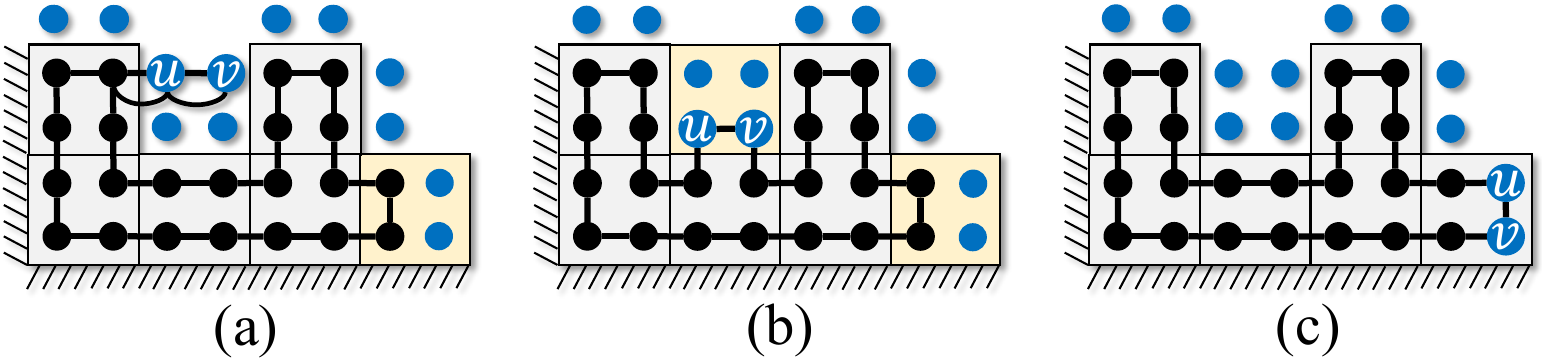}
\caption{An example of grow operator $o_g(i,e)$ and the coverage paths after it is executed. Blue circles and black circles represent vertices of the boundary vertex set $B_i$ and vertices of $V_{d,i}$, respectively. (a) An invalid grow operator without any parallel edge in $D_i$. (b)(c) Two valid grow operators.}
\label{fig:growing_op}
\end{figure}

\noindent\textbf{Grow Operators:}
A grow operator serves to expand a subgraph to cover an additional vertex that is already covered by other subgraphs.
Let $B_i$ denote the set of vertices that are not part of $D_i$ but adjacent to a vertex of $D_i$, i.e., $B_i=\{v\in V_{d,i}\,|\, \exists\,(u,v)\in E_{d,i}, u\in V_d/V_{d,i}\}$.
Formally, a \textit{valid grow operator} $o_g(i,e)$ for $D_i$ adds edge $e=(u, v)\in F_{i}$ with $u,v\in B_i$ and all relevant connecting edges into $D_i$, where $\exists\, (p, q)\in E_{d,i}$ such that $(u,p), (v,q)\in E_{d,i}$.
In essence, a valid grow operator can only add an edge $e=(u, v)$ iff there exists a parallel edge $(p, q)$ in $D_i$ (see Fig.~\ref{fig:growing_op}-(b)(c) for examples).
This design choice avoids introducing undesirable routing to the coverage path, as one case exemplified in Fig.~\ref{fig:growing_op}-(a).
After a grow operator $o_g(i,e)$ for $e=(u,v)$ is executed, vertices $u, v$ are added to $V^+$ since they are also covered by other subgraphs.

\begin{figure}[t]
\centering
\includegraphics[width=0.85\columnwidth]{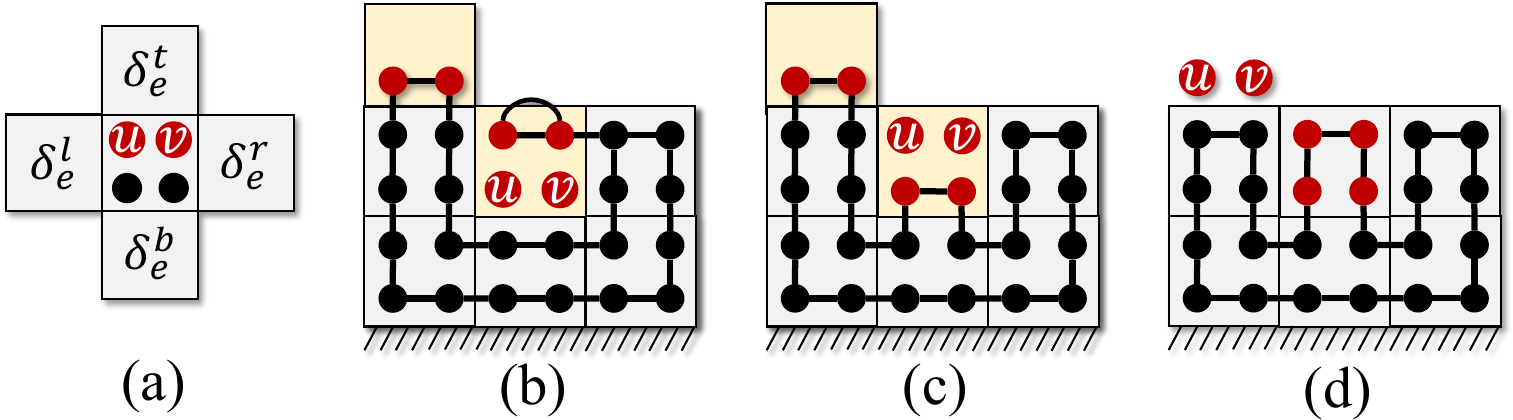}
\caption{An example of deduplicate operator $o_d(i,e)$. Red circles and black circles represent vertices in $V^+$ and not in $V^+$, respectively. (a) Four neighbors of $\delta_u$. (b) An invalid operator violating condition (3). (c)(d) Two valid operators.}
\label{fig:dedup_op}
\end{figure}

\noindent\textbf{Deduplicate operators:}
A deduplicate operator serves to remove unnecessary duplication from a subgraph.
For an edge $e=(u, v)\in F_{i}$, let $\delta^{\,t}_{e}$, $\delta^{\,b}_{e}$, $\delta^{\,l}_{e}$, and $\delta^{\,r}_{e}$ denote the four neighbors of $\delta_u$ (also equal to $\delta_v$) based on the positioning of $u$ and $v$.
Fig.~\ref{fig:dedup_op}-(a) demonstrates one example that can be rotated to various symmetric cases (with only $\delta^{\,t}_{e}$ potentially having its decomposed vertices adjacent to both $u$ and $v$).
Formally, a \textit{valid deduplicate operator} $o_d(i,e)$ for $D_i$ removes edge $e=(u, v)\in F_i$ with $u,v\in V^+\cap V_{d,i}$ and all relevant connecting edges from $D_i$ such that $D_i$ remains connected. This removal is constrained by three conditions if $\delta_u$ is complete: (1) $\delta_{e}^t$ is not in $V_{g,i}$; (2) all decomposed vertices of $\delta^{\,b}_{e}$ are in $V_{d,i}$; and (3) if $\delta=\delta_{e}^l,\delta_{e}^r$ is in $V_{g,i}$, then all decomposed vertices of both $\delta$ and its (only) common neighboring vertex with $\delta_{e}^b$ are in $V_{d,i}$. This design choice avoids introducing new duplication (such as the one shown in Fig.~\ref{fig:dedup_op}-(b)) to the coverage path when ESTC is applied to $D_i$. Otherwise, when $\delta_u$ is incomplete, the removal is not constrained by specific conditions. This design choice is based on the observation that the edge connecting an incomplete $\delta_u$ has low priority in the MST construction of ESTC. Consequently, $\delta_u$ is either positioned on or proximate to the periphery of the MST, and removing $e=(u, v)$ often does not introduce unnecessary or any duplication. 
We illustrate two valid deduplicate operators for an incomplete and complete $\delta_u$ in Fig.~\ref{fig:dedup_op}-(c) and Fig.~\ref{fig:dedup_op}-(d), respectively.
After a deduplicate operator $o_d(i,e)$ for $e=(u,v)$ is executed, vertices $u, v$ are removed from $V^+$ if they are not covered more than once by other subgraphs.

\noindent\textbf{Exchange Operators:}
An exchange operator combines a grow operator and a deduplicate operator so that the deduplicate operator immediately removes the new duplication introduced by the grow operator, potentially improving the effectiveness of a single operator and speeding up the convergence of the local search.
Formally, a \textit{valid exchange operator} $o_e(i,j,e)$ adds $e=(u, v)\in F_i$ and all relevant connecting edges into $D_i$ and removes them from $D_j$, where $o_g(i,e)$ and $o_d(j,e)$ are both valid, except that the condition $u,v\in V^+\cap V_{d,j}$ for $o_d(j,e)$ is now replaced by $u,v\in V_{d,j}$ since $u,v$ are not necessarily in $V^+$.
After an exchange operator $o_e(i,j,e)$ is executed, $V^+$ remains unchanged.

\noindent\textbf{Properties:} After any valid operator is executed, (1) $D_i$ remains connected since only deduplicate operators can remove edges but a valid deduplicate guarantees that $D_i$ remains connected, and (2) the property of the union of the vertices of all subgraphs being the vertex set $D$ is not affected since no vertex coverage is lost.

\section{LS-MCPP}\label{sec:ls}
In this section, we present LS-MCPP, a novel local search framework for effective MCPP solutions. LS-MCPP integrates the boundary editing operators, directly operating on the decomposed graph $D$ to explore good coverage paths by adjusting subgraphs $D_i$.
We categorize $D_i$ as a \textit{light subgraph} if its coverage path cost $c(\pi_i)$ from ESTC is no larger than the average coverage path cost $\bar{c}$ of all subgraphs; otherwise, it is a \textit{heavy subgraph}.
 LS-MCPP employs grow operators on light subgraphs and deduplicate operators on heavy subgraphs to eliminate redundancy and exchange operators to balance path costs between subgraphs. The goal is to achieve a cost-equilibrium solution with minimal duplication.
At a higher level, LS-MCPP employs a hierarchical sampling approach for efficient exploration of the constructed neighborhood in each iteration of its local search. It selects an operator pool using \textit{roulette wheel} selection~\cite{goldberg1989genetic} from three pools, each containing operators of the same type. Then, new heuristics are introduced to draw operators from the selected pool. LS-MCPP also calls a deduplication function periodically to exploit the current neighborhood and achieve a low-makespan solution.

\noindent\textbf{Pseudocode (Alg.~\ref{alg:ls-mcpp}):}
LS-MCPP takes an initial MCPP solution as input. It starts with initializing three pools $O_g, O_d, O_e$ of grow operators, deduplicate operators, and exchange operators, respectively (lines~\ref{alg:ls:init_Og}-\ref{alg:ls:init_Oe}), and initializing the temperature scalar $t$ and the pool weight vector $\mathbf{p}$ (line~\ref{alg:ls:init}).
The scalar $t$ originated from \textit{simulated annealing}~\cite{van1987simulated} is widely used to adaptively accept non-improving operators in search-based algorithms.
The vector $\mathbf{p}$ represents the pool weight for each of the three aforementioned pools and is thereby used for roulette wheel selection to select a pool $O$ from $\mathcal{O}$ in each iteration (line~\ref{alg:ls:rw_sampling}), where $\sigma(\mathbf{p})$ is the \textit{softmax} function of $\mathbf{p}$ determining the probability of selecting each pool.
Once the pool $O$ is selected, its corresponding pool weight $\mathbf{p}[O]$ is updated with a weight decay factor $\gamma\in[0,1]$ in line~\ref{alg:ls:update_pool_weight}.
Similar to the pool selection, LS-MCPP samples an operator $o$ from the selected pool $O$ with $\sigma(\mathbf{h})$ as the probability function (line~\ref{alg:ls:op_sampling}), where $\mathbf{h}$ is the vector of heuristic function $h$ evaluated on all operators in $O$ (line~\ref{alg:ls:heur_vec}), which we will describe in the next paragraph.
Then, LS-MCPP evaluates operator $o$ on its relating subgraph(s) and calculates the makespan increment $\Delta\tau$ after applying ESTC on the relating subgraph(s) (line~\ref{alg:ls:makespan_inc}) to determine whether to accept the new solution (i.e., applying $o$ and updating $\Pi$).
LS-MCPP accepts the new solution if it has a smaller makespan, and, otherwise, with a probability $\exp{(-\Delta\tau/t)}\in[0,1]$ (lines~\ref{alg:ls:ada_acc_st}-\ref{alg:ls:ada_acc_ed}).
Such an adaptive acceptance criterion is dynamically adjusted by $\Delta\tau$ divided by temperature $t$ that is scaled down by the decay factor $\alpha\in[0,1]$ for every iteration (line~\ref{alg:ls:update_pools}) so that it gets harder to accept a non-improving operator over iterations, which is the simulated annealing strategy to skip the local minimum.
LS-MCPP calls a function {\scshape forcedDeduplication()} (to be described later) every $S$ iteration or when the current solution has the smallest makespan (line~\ref{alg:ls:force_dedup}) and records the makespan-minimal solution in $\Pi^*$ (line~\ref{alg:ls:record_opt}).
If operator $o$ is applied thereby updating its relating subgraph(s), then the pools of $\mathcal{O}$ are updated with all the operator(s) related to the modified subgraph(s) (line~\ref{alg:ls:update_pools}). 
LS-MCPP terminates and returns the improved solution $\Pi^*$ when it reaches the maximum number of iterations $M$ (line~\ref{alg:ls:return}).

\begin{algorithm}[tb]
\DontPrintSemicolon
\linespread{0.5}\selectfont
\caption{LS-MCPP}\label{alg:ls-mcpp}
\SetKwInput{KwInput}{Input}
\SetKwInput{KwParam}{Param}
\KwInput{MCPP instance $(G, D, R)$, initial solution $\Pi$}
\KwParam{Max iteration $M\in\mathbb{Z}^+$, forced deduplication step $S\in\mathbb{Z}^+$, temperature decay factor $\alpha\in[0,1]$, pool weight decay factor $\gamma\in[0, 1]$}
$O_g\gets\bigcup_{i\in I}\{o_g(i,e)\,|\,c(\pi_i)\leq\bar{c}\}$ \;\label{alg:ls:init_Og}
$O_d\gets\bigcup_{i\in I}\{o_d(i,e)\,|\,c(\pi_i)>\bar{c}\}$ \;\label{alg:ls:init_Od}
$O_e\gets\bigcup_{i,j\in I}$$\{o_e(i,j,e)\,|\,c(\pi_i)\leq\bar{c}\}$ \;\label{alg:ls:init_Oe}
$\Pi^*\gets\Pi, t\gets 1, \mathbf{p}\gets [1, 1, 1]^T, \mathcal{O}=\{O_g, O_d, O_e\}$ \;\label{alg:ls:init}
\For{$i\gets 1, 2,..., M$}
{
    $O\sim\mathcal{O}$ w/ probability function $P(\mathcal{O})=\sigma(\mathbf{p})$ \;\label{alg:ls:rw_sampling}
    $\mathbf{p}[O]\gets (1-\gamma)\cdot\mathbf{p}[O]+\gamma\cdot\max (-\Delta\tau, 0)$ \;\label{alg:ls:update_pool_weight}
    $\mathbf{h}\gets$ the heuristic value vector $[h(o)]^T_{o\in O}$ \;\label{alg:ls:heur_vec}
    $o\sim O$ w/ probability function $P(O)=\sigma(\mathbf{h})$ \;\label{alg:ls:op_sampling}
    $\Delta\tau\gets$ makespan increment after using ESTC on updated subgraph(s) with $o$\;\label{alg:ls:makespan_inc}
    \eIf{$\Delta\tau<0$}{\label{alg:ls:ada_acc_st}
        Apply operator $o$ and update $\Pi$ \;
    }
    {
        \scalebox{0.95}{Apply $o$ and update $\Pi$ w/ probability of $\exp{(\frac{-\Delta\tau}{t})}$} \;\label{alg:ls:ada_acc_ed}
    }
    \If{$i\,\%\, S = 0 \vee \Delta\tau < 0$}
    {
        \scshape{forcedDeduplication}($\Pi$, $O_d$) \;\label{alg:ls:force_dedup}
    }
    Assign $\Pi$ to $\Pi^*$ if $\Pi$ is better than $\Pi^*$ \;\label{alg:ls:record_opt}
    Update each pool of $\mathcal{O}$ and assign $t$ to $\alpha\cdot t$\;\label{alg:ls:update_pools}
}
\Return Improved MCPP solution $\Pi^*$ \;\label{alg:ls:return}
\vspace{3pt}
\renewcommand{\texttt}[1]{\scshape{\textsc{#1}}}
\SetKwFunction{FMain}{forcedDeduplication}
\SetKwProg{Fn}{Function}{:}{}
\Fn{\FMain{$\Pi$, $O_d$}}
{\label{func:forcededup}

\For{$i\in \argsort\left(\{c(\pi_i)\}_{i\in I}\right)$}
{\label{func:forcededup:main_loop_1}
    \While{$(u, v)\gets$ any U-turn in $\pi_i$}
        {\label{func:forcededup:find_U_turn}
            Remove $u, v$ from $D_i$ and update $\pi_i$ \;\label{func:forcededup:update}
        }
}

\For{$i\in \argsort\left(\{c(\pi_i)\}_{i\in I}\right)$}
{\label{func:forcededup:main_loop_2}
    \While{$\{o_d(j,e)\in O_d\,|\,i=j\}\neq\emptyset$}
    {
        $o\gets \argmin\left([h(o)]^T_{o\in\{o_d(j,e)\in O_d\,|\,i=j\}}\right)$\;\label{func:forcededup:order}
        Apply operator $o$, update $\Pi$ and $O_d$\;\label{func:forcededup:update2}
    }
}
}

\end{algorithm}

\noindent\textbf{Operator Sampling with Heuristics:}
LS-MCPP should carefully determine the probability of selecting each operator to better explore the neighborhood when sampling an operator from the selected pool (line~\ref{alg:ls:op_sampling} of Alg.~\ref{alg:ls-mcpp}).
We propose three heuristic functions tailored to the three types of operators to evaluate their potential in guiding the neighborhood search and improving the solution.
An operator with a larger heuristic value results in a higher probability of being sampled.
The heuristic function for grow operators has two considerations: (1) prioritizing growing light subgraphs with small coverage path costs and (2) prioritizing covering vertices with less duplication.
Formally, the heuristic value for a grow operator $o_g(i,e)$ with edge $e=(u,v)$ is defined as $h(o_g) = -k\cdot c(\pi_i) - (n_u + n_v)/2$.
Note that $k \geq n_v$ holds for any $v\in V_d$, thereby prioritizing consideration (1) (i.e., $c(\pi_i)$) over consideration (2) (i.e., $(n_u+n_v)/2$).
The heuristic function for deduplicate operators $o_d(i,e)$ is defined as the exact opposite of the one for grow operators: $h(o_d) = k\cdot c(\pi_i) + (n_u + n_v)/2$.
The heuristic function for exchange operators $o_e(i,j,e)$ is defined as $h(o_e) = c(\pi_j) - c(\pi_i)$ to prioritize two subgraphs with a big difference in their coverage path costs.

\begin{figure}[t]
\centering
\includegraphics[width=0.7\columnwidth]{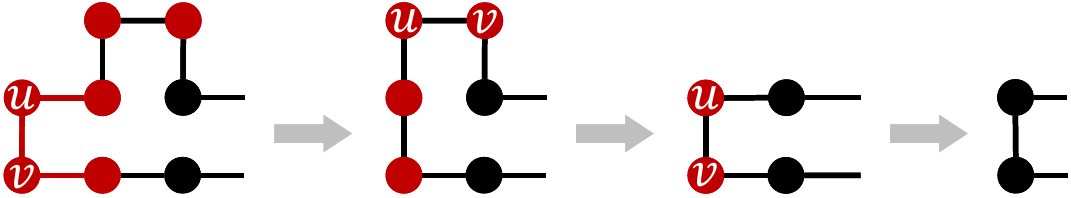} 
\caption{{\scshape forcedDeduplication()} executed on a coverage path $\pi_i$. Each red circle represents a duplication of $V_{d,i}$. In each frame, a U-turn $(u, v)\in\pi_i$ is found and $u, v$ are removed from $\pi_i$ until no U-turn exists.}
\label{fig:U_turn}
\end{figure}

\noindent\textbf{Forced Deduplication:}\label{subsec:force_dedup}
A key consideration towards the success of LS-MCPP is to restrict duplication to only those necessarily needed over the neighborhood construction; for instance, a narrow path connecting two separate regions of the terrain in the \textit{flr-s} instance~\cite{tang2023mixed}.
For this purpose, {\scshape forcedDeduplication()} (line~\ref{func:forcededup} of Alg.~\ref{alg:ls-mcpp}) aims to deduplicate all possible duplication in two folds.
The first part (lines~\ref{func:forcededup:main_loop_1}-\ref{func:forcededup:update}) iterates through each coverage path $\pi_i\in\Pi$ in a cost-decreasing order to remove any U-turn in it (lines~\ref{func:forcededup:find_U_turn}-\ref{func:forcededup:update}).
A U-turn is an edge $(u, v)\in \pi_i$ with $u, v\in V^+\cap V_{d,i}$, satisfying $\exists\,(p, q)\in\pi_i$ such that $(u, p), (v,q) \in\pi_i$. (so that $p\rightarrow u\rightarrow v\rightarrow q$ forms a ``U-turn''.)
The second part (lines~\ref{func:forcededup:main_loop_2}-\ref{func:forcededup:update2}) tries to recursively apply all valid deduplicate operators in a descending order of the coverage path costs (line~\ref{func:forcededup:main_loop_2}) and an ascending order of the heuristic values (line~\ref{func:forcededup:order}).
Fig.~\ref{fig:U_turn} demonstrates {\scshape forcedDeduplication()} on a coverage path.
It is worth noting that the first part can remove edges that do not constitute valid deduplicate operators, and some valid deduplicate operators can remove edges that are not U-turns (depending on the spanning tree generated by ESTC).
The complementary relation between the two parts yields a better performance of the proposed LS-MCPP.

\begin{table}[t]
\centering
\renewcommand{\arraystretch}{0.85}
\fontsize{15}{18}\selectfont\scalebox{0.59}{%
\begin{tabular}{|c|c|c|c|c|c|}
\hline
\textbf{Instance} & \textbf{Grid spec.} & \bm{$\lvert V_g \rvert$} & 
\bm{$\lvert E_g \rvert$} & \bm{$k$} & \textbf{weighted} \\\hline
\textit{AR0701SR} & 107$\times$117 & 4,860 & 8,581 & 20 & $\checkmark$ \\\hline
\textit{Shanghai2} & 128$\times$128 & 11,793 & 22,311 & 25 & $\times$ \\\hline
\textit{NewYork1} & 128$\times$128 & 11,892 & 21,954 & 32 & $\checkmark$ \\\hline
\end{tabular}
}%
\caption{MCPP instance specification. 
}
\label{tab:list_istc}
\end{table}

\section{Empirical Evaluation}
\begin{figure}[t]
\centering
\includegraphics[width=0.75\columnwidth]{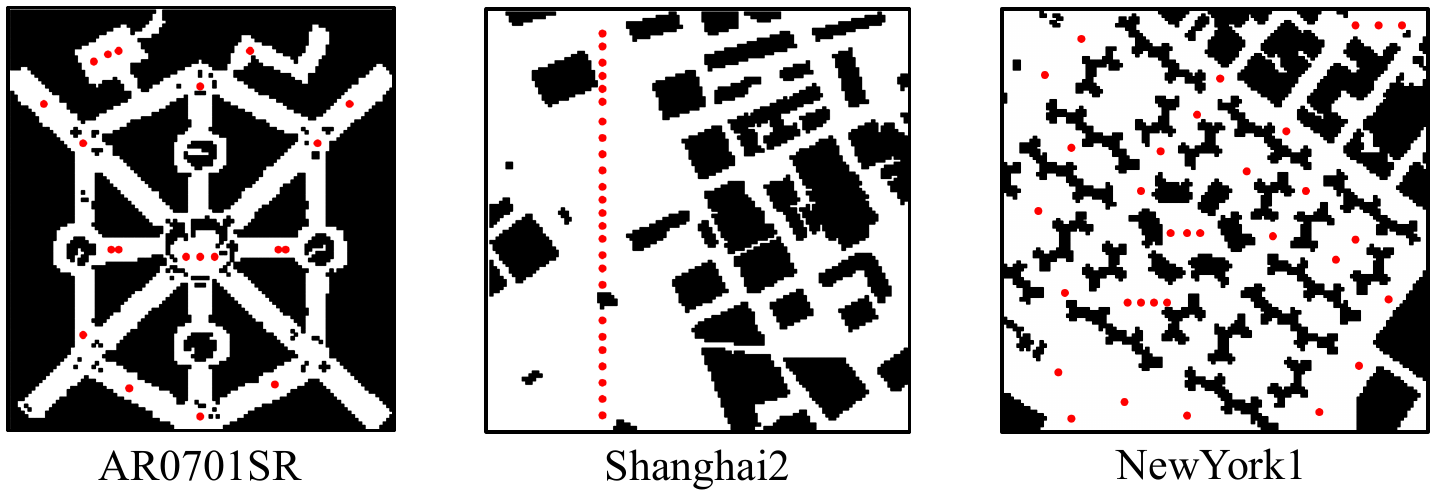} 
\caption{Three very large-scale MCPP instances adopted from 2D pathfinding benmarks~\cite{sturtevant2012benchmarks}. Red circles represent the initial vertices of robots.}
\label{fig:istc}
\end{figure}

This section describes our experimental results on an \textit{Intel}\textsuperscript{\textregistered} \textit{Xeon}\textsuperscript{\textregistered} Gold 5218 2.30 GHz Linux server with 300 GB memory. All LS-MCPP experiments were conducted across 12 repetitions with seeds spanning from 0 to 11. 
\textbf{Instances:}
We use MCPP instances from~\cite{tang2023mixed}, where their numbers of graph vertices, graph edges, and robots range from $46$ to $824$, $60$ to $1495$, and $4$ to $12$, respectively.
As shown in Fig.~\ref{fig:istc} and Tab.~\ref{tab:list_istc}, we design three additional larger instances whose terrain graphs are adopted from popular 2D pathfinding benchmarks~\cite{sturtevant2012benchmarks}. All the above instances have complete terrain graphs.
\textbf{Baselines:}
We use MFC~\cite{zheng2010multirobot}, MSTC$^*$~\cite{tang2021mstc}, and MIP/MIP(H)~\cite{tang2023mixed} as baseline algorithms that all use STC to transform a tree cover on $G$ to an MCPP solution. Both MFC and MSTC$^*$ compute a suboptimal tree cover. MIP computes an optimal tree cover with a MIP model. MIP(H) computes a suboptimal tree cover with a reduced-size MIP model but is more efficient than MIP.
We also design a new baseline VOR that first computes the Voronoi diagram (i.e., $k$ subgraphs) of $G$ based on the initial vertices of robots and then uses ESTC to generate coverage paths on the subgraphs.
\textbf{Parameters:}
For all instances solved with LS-MCPP, the pool weight decay factor $\gamma$ is set to $0.01$, and the temperature decay factor $\alpha$ is set to $\exp{\frac{\log{0.2}}{M}}$ to decrease the temperature $t$ from $1$ to $0.2$. 
For instances from~\cite{tang2023mixed}, we set the max iteration $M$ to $3e3$ and the forced deduplication step $S$ to $1e2$,
For the three additional larger instances in Tab.~\ref{tab:list_istc}, we set $M$ to $1.5e4$ and $S$ to $5e2$.

\begin{figure}[t]
\centering
\includegraphics[width=0.85\columnwidth]{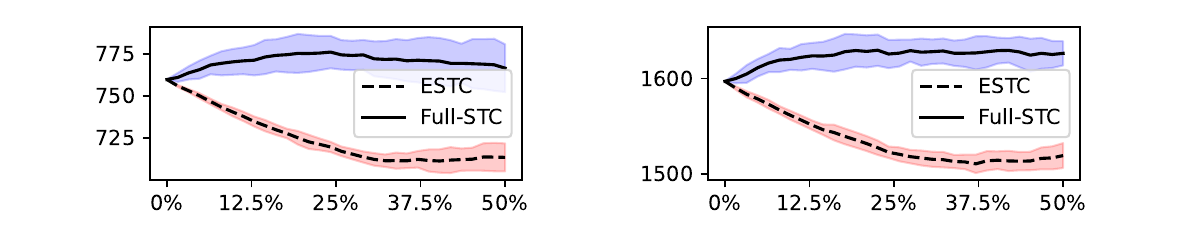} 
\caption{Single-robot coverage path costs (y-axis) with the percentages of incomplete terrain vertices (x-axis). Left and right sub-figures correspond to \textit{flr-l} and \textit{trn1-l}, respectively.}
\label{fig:exp_estc}
\end{figure}

\subsection{Ablation Study}\label{subsec:ablation}
We use two instances \textit{flr-l} and \textit{trn1-l} to conduct an ablation study on different components of LS-MCPP.

\noindent\textbf{ESTC vs Full-STC:}
For each instance, we make its terrain graph incomplete by randomly sampling a set $V'_g\subseteq V_g$ and removing the respective 1, 2`, or 3 decomposed vertices from each $\delta\in V'_g$ randomly, without making the decomposed graph unconnected.
Fig.~\ref{fig:exp_estc} shows that ESTC consistently outperforms Full-STC on incomplete terrain graphs.

\begin{figure}[t]
\centering
\includegraphics[width=0.98\columnwidth]{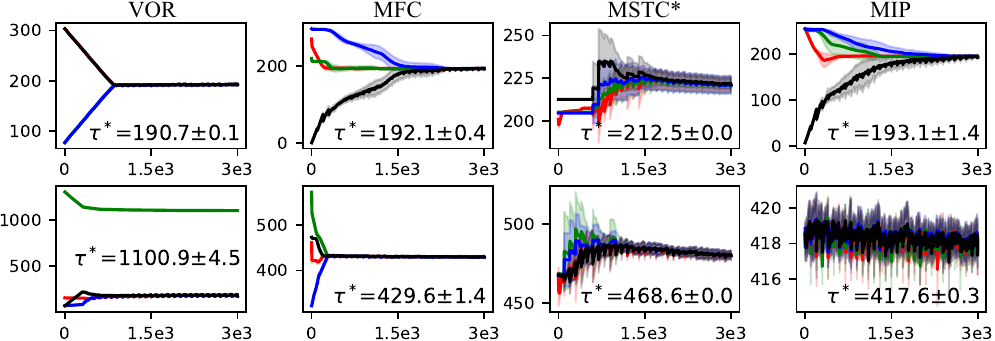} 
\caption{Path costs (y-axis) of 4 robots for LS-MCPP over iterations (x-axis) with different initial solutions. The upper and lower rows correspond to \textit{flr-l} and \textit{trn1-l}, respectively.}
\label{fig:exp_init_sol}
\end{figure}

\noindent\textbf{Initial Solution:}
We compare the performance of LS-MCPP with different initial solutions in Fig.~\ref{fig:exp_init_sol}.
LS-MCPP shows convergence for all cases, except for \textit{trn1-l} with VOR as a result of its low-quality solution due to its dependence on the initial vertices of robots.
MSTC$^*$ tends to require more search iterations than MFC and MIP before reaching a good solution since it partitions a single MST to obtain a tree cover that often results in a different solution structure from the direct tree cover computation employed by MFC and MIP.
As MIP might yield near-optimal solutions (e.g., \textit{trn1-l}) at the cost of longer runtime, the potential for improvement using LS-MCPP is limited.
Therefore, we select MFC to produce initial solutions for LS-MCPP, striking a balance between efficiency and solution quality.

\begin{figure}[t]
\centering
\includegraphics[width=0.98\columnwidth]{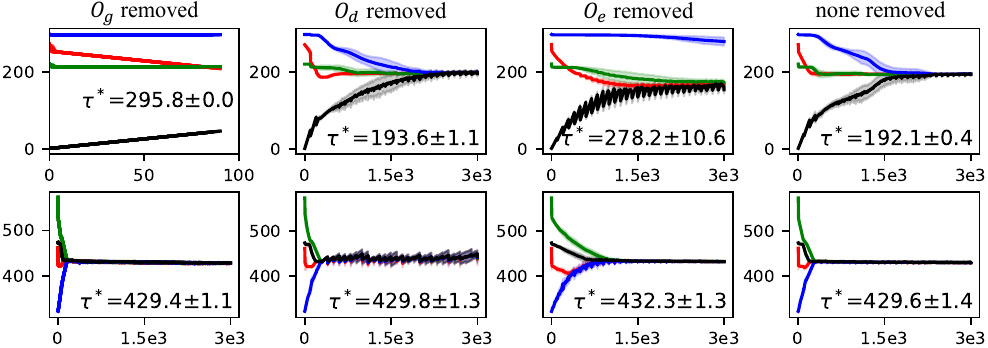} 
\caption{Path costs (y-axis) of 4 robots for LS-MCPP over iterations (x-axis) with different operators removed. The upper and lower row correspond to \textit{flr-l} and \textit{trn1-l}, respectively.}
\label{fig:exp_operators}
\end{figure}

\noindent\textbf{Boundary Editing Operators:}
We assess the impact of different types of operators on LS-MCPP in Fig.~\ref{fig:exp_operators}. 
The first column removes grow operators, causing an early stop of LS-MCPP before the $3e3$ iteration limit. This can limit exploration of the solution space, especially when exchange operators struggle to grow light subgraphs effectively due to the graph structure (e.g., \textit{flr-l} instance).
The second column removes deduplicate operators, leading to instability and worse performance. These operators are crucial for reduplication between two {\scshape ForcedDeduplication()} calls.
The third column removes exchange operators, hindering LS-MCPP convergence and solution quality improvement.

\begin{figure}[t]
\centering
\includegraphics[width=0.98\columnwidth]{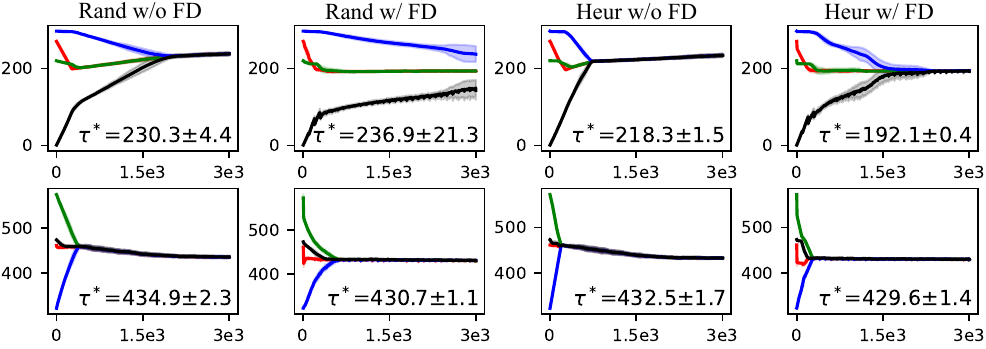} 
\caption{Path costs (y-axis) of 4 robots for LS-MCPP over iterations (x-axis) with different operator sampling and whether {\scshape ForcedDeduplication} (FD) is used. The upper and lower row correspond to \textit{flr-l} and \textit{trn1-l}, respectively.}
\label{fig:exp_sampling_fd}
\end{figure}

\noindent\textbf{Operator Sampling:}
We use a uniform random operator sampling method (Rand) to compare with the proposed heuristic-based operator sampling method (Heur).
In Fig.~\ref{fig:exp_sampling_fd}, we compare the first and third columns, and the second and fourth columns.
We can see that Heur is much better than Rand in \textit{flr-l}; but it is only slightly better than Rand in \textit{trn1-l}, possibly because \textit{trn1-l} is a relatively simpler instance for LS-MCPP thus the difference between the two sampling methods is trivial.

\noindent\textbf{Forced Deduplication:}
In Fig.~\ref{fig:exp_sampling_fd}, we validate the function {\scshape ForcedDeduplication} (FD) of LS-MCPP by comparing the first and second columns, and the third and fourth columns.
The periodic calling of FD makes the subgraphs return to states with only necessary duplications remaining, guiding LS-MCPP to better solutions.

\subsection{Performance Comparison}
We compare LS-MCPP with the baselines in Tab.~\ref{tab:res_comp} with a runtime limit of 24 hours.
In summary, LS-MCPP outperforms VOR, MFC, MSTC$^*$ for all instances within 20 minutes, demonstrating a makespan reduction of up to 67.0\%, 35.7\%, and 30.3\%, and on average, 50.4\%, 26.7\%, and 13.4\%, respectively.
For the first six smaller instances where MIP(H) and MIP can almost compute the optimal tree cover, LS-MCPP achieves an average makepan reduction of -1.02\% and 1.16\% with orders of magnitude faster runtime.
Furthermore, we use the same method as in the ESTC ablation study to make 20\% of terrain vertices incomplete for each instance. The last column of Tab.~\ref{tab:res_comp} reports results for such instances where we generate covering subtrees of the terrain graph using RTC~\cite{even2004min} (used internally by MFC) followed by ESTC on the inducing subgraphs of the RTC subtrees to obtain the initial solution for LS-MCPP.
This is necessary since no baseline can handle incomplete terrain graphs. We observe that removing decomposed vertices can alter the connectivity of its neighboring vertices, thus affecting the overall structure of both the decomposed graph and the augmented terrain graph, resulting in a worse LS-MCPP solution than that of the original instance.

\begin{table}[t]
\centering
\setlength\tabcolsep{0.7pt}
\renewcommand{\arraystretch}{0.9}
\fontsize{15}{18}\selectfont\scalebox{0.595}{%
\begin{tabular}{|c|c|c|c|c|c|c|c|c|c|}
\hline
                               & \textbf{VOR}  & \textbf{MFC} & \textbf{MSTC$^*$} & \multicolumn{2}{c|}{\textbf{MIP(H)}} & \multicolumn{2}{c|}{\textbf{MIP}} & \multicolumn{2}{c|}{\textbf{LS-MCPP (Ours)}} \\ \hline
\multirow{2}{*}{\shortstack{\textit{flr-s}}} & 42.75 & 23 & 21 & \textbf{16} & \multirow{2}{*}{\rotatebox[origin=c]{270}{0.0\%}} & \textbf{16} & \multirow{2}{*}{\rotatebox[origin=c]{270}{0.0\%}} & 16.75$\pm$0.3 & 21.5$\pm$2.1 \\ \cline{2-5}\cline{7-7}\cline{9-10} 
                                & \textbf{0.01s} & 0.03s & 0.02s & 13s & & 20s & & 5.3$\pm$0.13s & 3.6$\pm$2.9s            \\ \hline
\multirow{2}{*}{\textit{trn-m}} & 420.2 & 368.2 & 269.5 & 246.7 & \multirow{2}{*}{\rotatebox[origin=c]{270}{0.6\%}} & 246.7 & \multirow{2}{*}{\rotatebox[origin=c]{270}{1.6\%}} & \textbf{244.5$\pm$0.8} & 245$\pm$12.2 \\ \cline{2-5}\cline{7-7}\cline{9-10}  
                                & \textbf{0.04s} & 0.58s & 0.32s & 24h & & 24h & & 17.8$\pm$0.1s & 19.6$\pm$1.3s\\ \hline
\multirow{2}{*}{\textit{mze-m}} & 134.8 & 67 & 69 & 52 & \multirow{2}{*}{\rotatebox[origin=c]{270}{0.0\%}} & \textbf{51} & \multirow{2}{*}{\rotatebox[origin=c]{270}{20\%}} & 54.3$\pm$2.9 & 65.0$\pm$4.2 \\ \cline{2-5}\cline{7-7}\cline{9-10}  
                                & \textbf{0.02s} & 0.35s & 0.34s & 4.9s & & 24h & & 10.7$\pm$0.1s & 9.0$\pm$2.9s\\ \hline
\multirow{2}{*}{\textit{flr-l}} & 303.8 & 294 & 212.5 & 207 & \multirow{2}{*}{\rotatebox[origin=c]{270}{8.3\%}} & 254 & \multirow{2}{*}{\rotatebox[origin=c]{270}{25\%}} & \textbf{192.1$\pm$0.4} & 207$\pm$20.2      \\ \cline{2-5}\cline{7-7}\cline{9-10}  
                                & \textbf{0.07s} & 2.37s & 0.09s & 24h & & 24h & & 30.5$\pm$1.3s & 35.2$\pm$1.6s\\ \hline
\multirow{2}{*}{\textit{mze-l}} & 193.8 & 105 & 139.5 & \textbf{91.5} & \multirow{2}{*}{\rotatebox[origin=c]{270}{0.0\%}} & 93 & \multirow{2}{*}{\rotatebox[origin=c]{270}{25\%}} & 97.3$\pm$0.5 & 104.8$\pm$7.6\\ \cline{2-5}\cline{7-7}\cline{9-10}  
                                & \textbf{0.04s} & 0.10s & 0.58s & 54s & & 24h & & 15.4$\pm$0.2s & 12.9$\pm$5.8s\\ \hline
\multirow{2}{*}{\textit{trn1-l}} & 1,303 & 597.4 & 468.6 & 435.1 & \multirow{2}{*}{\rotatebox[origin=c]{270}{9.7\%}} & \textbf{419.1} & \multirow{2}{*}{\rotatebox[origin=c]{270}{6.1\%}} & 429.6$\pm$1.4 & 444$\pm$11.2\\ \cline{2-5}\cline{7-7}\cline{9-10}  
                                & \textbf{0.07s} & 1.67s & 1.93s & 24h & & 24h & & 34.3$\pm$0.4s & 37.3$\pm$1.6s\\ \hline
\multirow{2}{*}{\shortstack{\textit{AR07}\\\textit{01SR}}} & 1,301  & 1,254 & 878.7 & 1,351 & \multirow{2}{*}{\rotatebox[origin=c]{270}{42\%}}& 1,333 & \multirow{2}{*}{\rotatebox[origin=c]{270}{55\%}} & \textbf{805.9$\pm$7.9} & 1,024$\pm$74      \\ \cline{2-5}\cline{7-7}\cline{9-10}  
                                & \textbf{0.47s} & 36.9s & 9.91s & 24h & & 24h & & 6.3$\pm$0.1m & 6.2$\pm$0.3m\\ \hline
\multirow{2}{*}{\shortstack{\textit{Shan}\\\textit{ghai2}}} & 1,451 & 754 & 576.5 & 1,509 & \multirow{2}{*}{\rotatebox[origin=c]{270}{59\%}} & / & \multirow{2}{*}{/} & \textbf{570.9$\pm$2.1} & 719$\pm$73.0      \\ \cline{2-5}\cline{7-7}\cline{9-10}  
                                & \textbf{1.21s} & 7.2m & 2.21s & 24h & & / & & 17.6$\pm$2.1m & 12.5$\pm$0.8m\\ \hline
\multirow{2}{*}{\shortstack{\textit{New}\\\textit{York1}}} & 1,735  & 1,530  & 1,208 & 4,736 & \multirow{2}{*}{\rotatebox[origin=c]{270}{80\%}} & / & \multirow{2}{*}{/} & \textbf{1,062$\pm$14} & 1,563$\pm$110      \\ \cline{2-5}\cline{7-7}\cline{9-10}  
& \textbf{1.13s} & 2.4m & 20.2s & 24h & & / & & 15.0$\pm$4.1m & 12.6$\pm$0.5m\\ \hline
\end{tabular}
}%
\caption{Performance comparison. Each instance-method cell reports the makespan at the top and the runtime at the bottom. The percentages for MIP(H)/MIP are the gap between the solution and the lower bound returned by the MIP solver. The last column is for incomplete terrain graphs.}\label{tab:res_comp}
\end{table}

\section{Conclusions and Future Work}

We introduced LS-MCPP, a new local search framework that integrates the novel standalone ESTC paradigm and boundary editing operators to systematically explore MCPP solutions directly on decomposed graphs for the first time.
Future work includes enhancing operator selection with machine learning techniques, speeding up LS-MCPP for larger-scale instances with parallelization techniques, and addressing potential inter-robot collisions in occluded environments with multi-agent pathfinding \cite{stern2019multi} techniques.

\section*{Acknowledgements}
This work was supported by the NSERC under grant number RGPIN2020-06540 and a CFI JELF award.

\bibliography{ref}

\end{document}